\documentclass{article}

\PassOptionsToPackage{numbers,compress}{natbib}
\usepackage[main,final]{neurips_2026}

\usepackage[utf8]{inputenc}
\usepackage[T1]{fontenc}
\usepackage{hyperref}
\usepackage{url}
\usepackage{booktabs}
\usepackage{amsfonts}
\usepackage{nicefrac}
\usepackage{microtype}
\usepackage{xcolor}
\usepackage{graphicx}
\usepackage{subcaption}
\usepackage{amsmath}
\usepackage{amssymb}
\usepackage{mathtools}
\usepackage[capitalize,noabbrev]{cleveref}

\title{Beyond Spatio-Temporal Priors: A Generalizable Approach for Dense Correspondence Matching}
\author{Luping Liu\textsuperscript{1,2}\thanks{Work done while Luping Liu and Yifan Wang were interns at ByteDance Seed.} \quad
  Bingyi Kang\textsuperscript{2} \quad
  Yifan Wang\textsuperscript{2,3}\footnotemark[1] \quad
  Dong Xu\textsuperscript{1}\thanks{Corresponding author.} \\
  \normalfont\textsuperscript{1}The University of Hong Kong \quad
  \textsuperscript{2}ByteDance Seed \\
  \normalfont\textsuperscript{3}Zhejiang University \\
  \normalfont\texttt{luping.liu@connect.hku.hk}\quad\texttt{bingykang@gmail.com} \\
  \normalfont\texttt{accwyf@gmail.com}\quad\texttt{dongxu@cs.hku.hk}
}

\begin{document}

\maketitle
\suppressfloats[t]

\begin{abstract}

   Dense correspondence matching has historically been bounded by simplifying spatio-temporal priors, such as smooth motion and rigid geometry. While effective for classical tasks, these assumptions break down in image editing and reference-guided generation (IEG), where transformations can preserve visual identity while breaking physical continuity. To establish identity-preserving correspondence across such transformations, we introduce FreeMatching, a generalizable framework combining generative and semantic foundation representations with heterogeneous supervision from classical datasets, tracked videos, and synthetic scenes. Teacher-guided iterative refinement further improves correspondence in IEG without dense correspondence annotations. Experimentally, a single FreeMatching model substantially improves correspondence quality on challenging IEG image pairs while retaining competitive performance on classical benchmarks. Furthermore, we demonstrate its utility as a quantitative metric for evaluating identity preservation, with scores that correlate with human judgment. The code is available at \url{https://github.com/luping-liu/FreeMatching}.
\end{abstract}

\begin{figure*}[t]
  \centering
  \includegraphics[width=0.88\linewidth]{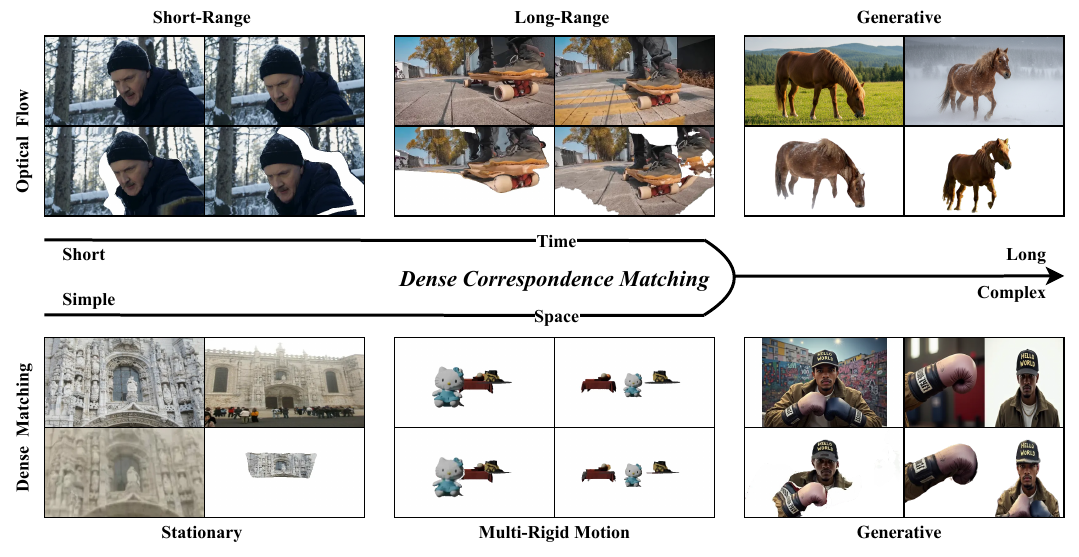}
  \captionof{figure}{Extending Dense Correspondence to IEG. FreeMatching matches across optical flow, geometric matching, and IEG. Each block shows input pairs above and our bidirectional warps below. The classical examples vary in temporal separation (top) and geometric complexity (bottom).}
  \label{fig:task_definition}
\end{figure*}

\section{Introduction}

Dense correspondence matching has historically been fractured into distinct paradigms: optical flow~\cite{horn1981determining, lucas1981iterative, teed2020raft} for temporal motion and dense matching~\cite{barnes2009patchmatch, edstedt2024roma} for geometric views. While unified frameworks like UniMatch~\cite{xu2023unifying} and UFM~\cite{zhang2025ufm} have bridged these tasks, they remain constrained by simplifying spatio-temporal priors. As illustrated in \cref{fig:task_definition}, correspondence exists on a continuous spectrum. While existing methods excel at the left end governed by smooth motion and rigid geometry, they falter at the frontier of image editing and reference-guided generation (IEG)~\cite{wu2025less, wu2025qwen, seedream2025seedream}. For example, \cref{fig:task_definition} shows a horse in a grassy landscape and a snowy scene, with substantial changes in pose and appearance while its identity remains recognizable. Such transformations can preserve visual identity while breaking physical continuity. Classical spatio-temporal assumptions no longer suffice, and existing methods can produce omissions or distortions, as shown in \cref{fig:baseline_visual}.

To move beyond restrictive spatio-temporal priors, we draw inspiration from human cognition~\cite{kahneman1992reviewing}. Human perception can recognize the same instance despite substantial changes in appearance and configuration. Motivated by this principle, we aim to learn identity-preserving~\cite{tumanyan2023plug} correspondence across the diverse tasks shown in \cref{fig:task_definition}.

To achieve this goal, we introduce \textit{FreeMatching}, which combines foundation-model representations, heterogeneous correspondence supervision, and teacher-guided iterative refinement. Specifically, we utilize FLUX2-4B~\cite{bfl2026flux2klein} and DINOv3~\cite{simeoni2025dinov3} to inherit rich generative and semantic knowledge, respectively. We develop this capability through two stages of training: (1) supervised learning on new large-scale tracked video~\cite{karaev2024cotracker, harley2025alltracker} and synthetic datasets, combined with classical datasets~\cite{dai2017scannet, mayer2016large}, where Huber loss~\cite{huber1992robust} is employed to mitigate annotation noise; and (2) teacher-guided iterative refinement on IEG datasets without dense correspondence annotations. This stage refines local correspondences and provides supervision for covisibility. Together, these components extend correspondence learning beyond classical geometric and motion settings.

Our experiments demonstrate the power of FreeMatching's generalized approach. A single model achieves performance competitive with SOTA methods on classical optical flow and dense matching benchmarks, while substantially improving correspondence quality on the challenging IEG task. This success enables an application: using FreeMatching as a perceptual metric for quantifying identity preservation~\cite{huang2024vbench}, providing objective scores that correlate with human judgment.

Our contributions are threefold:
\begin{enumerate}
    \item We introduce FreeMatching, a generalizable framework that extends dense correspondence from classical geometric and motion settings to identity-preserving matching in IEG.
    \item We combine foundation-model representations, heterogeneous correspondence supervision, and teacher-guided iterative refinement through a two-stage training paradigm, supported by new large-scale video and synthetic datasets.
    \item We demonstrate competitive performance on classical tasks and substantial gains on IEG, enabling spatially resolved assessment of identity preservation with scores that correlate with human judgment.
\end{enumerate}

\section{Related Work}
\label{sec:related_work}

\subsection{Classical and Unified Correspondence}

The pursuit of dense correspondence has evolved significantly. Initially divided into optical flow for temporal motion~\cite{horn1981determining} and geometric matching for stationary scenes~\cite{barnes2009patchmatch}, the field was revolutionized by deep learning. Architectures like RAFT~\cite{teed2020raft} set a new standard with the paradigm of iterative refinement over cost volumes. This paradigm was pushed further by specialized models, such as GMFlow~\cite{xu2022gmflow} which framed flow as a global matching problem, and RoMa~\cite{edstedt2024roma}, which leverages powerful foundation models like DINOv2~\cite{oquab2023dinov2} for state-of-the-art wide-baseline matching.

This progress spurred a move towards unifying these tasks. Frameworks like UniMatch~\cite{xu2023unifying} and UFM~\cite{zhang2025ufm} have successfully demonstrated that a single, powerful model can match specialized methods on traditional benchmarks.

\subsection{Semantic Correspondence and Foundation Features}

Semantic correspondence research addresses geometry and viewpoint sensitivity through geometry-aware matching~\cite{zhang2024telling} and viewpoint-guided spherical maps~\cite{mariotti2024spherical}. GECO~\cite{hartwig2025geco} and diffusion-feature distillation~\cite{fundel2025distillation} study efficient correspondence representations, while Jamais Vu~\cite{mariotti2025jamais} examines the generalization gap of supervised semantic matching. Geometry-grounded models such as DUSt3R~\cite{wang2024dust3r} and VGGT~\cite{wang2025vggt} provide another source of foundation features, with recent work adapting VGGT priors to dense semantic matching~\cite{yang2026geometry}. These advances motivate learning representations that transfer across transformations.

\subsection{Learning Paradigms for Correspondence}

Training paradigms have co-evolved alongside architectures. The standard of supervised learning on synthetic datasets like FlyingThings~\cite{mayer2016large} has been augmented by two key trends: unsupervised learning through data distillation~\cite{liu2019ddflow}, and the unification of diverse data sources---such as pre-training optical flow models on wide-baseline data---to improve generalization~\cite{wang2024sea}.

Building on these advances, FreeMatching extends a single correspondence model from classical geometric and motion settings to IEG. We combine generative and semantic foundation representations with heterogeneous supervision from classical datasets, tracked videos, and synthetic scenes, followed by teacher-guided iterative refinement on IEG image pairs. This combination supports instance-level matching across changes in appearance, pose, and scene composition while retaining classical correspondence capabilities.

\section{FreeMatching}

We present \textit{FreeMatching}, a unified framework that extends classical correspondence matching to IEG. We first define identity-preserving correspondence, then describe the foundation-model architecture, supervised pre-training, and teacher-guided iterative refinement. Finally, we introduce a reconstruction-based evaluation protocol for matching quality and identity-preserving consistency.

\subsection{Task Definition}

Classical correspondence tasks use geometric and temporal structure to establish matches. IEG extends this setting: an instance can retain its visual identity even when changes in appearance, pose, and scene composition break physical continuity.

We formulate dense correspondence as identity-preserving visual matching. Rather than requiring correspondences to follow rigid geometry or smooth motion, we seek to match regions depicting the same underlying instance across changes in viewpoint, pose, appearance, and scene composition. This formulation encompasses classical correspondence tasks and extends to IEG, wherever an identifiable visual counterpart remains.

Formally, given a source image $I_a$ and a target image $I_b \in \mathbb{R}^{H \times W \times 3}$, our goal is to predict a dense backward correspondence map $C \in \mathbb{R}^{H \times W \times 2}$ and a covisibility mask $M \in [0,1]^{H \times W}$. Specifically, $C(x, y) = (u, v)$ denotes that pixel $(x, y)$ in $I_b$ corresponds to the absolute coordinate $(u, v)$ in $I_a$. The mask indicates whether a valid counterpart exists in the source image. Target regions without an identifiable source counterpart, such as newly synthesized content or content revealed after an occluder is removed, have a covisibility target of $M=0$. Shared regions remain matchable despite changes in appearance or configuration.

\subsection{Foundation Models}

\begin{figure}
    \centering
    \includegraphics[width=0.70\linewidth]{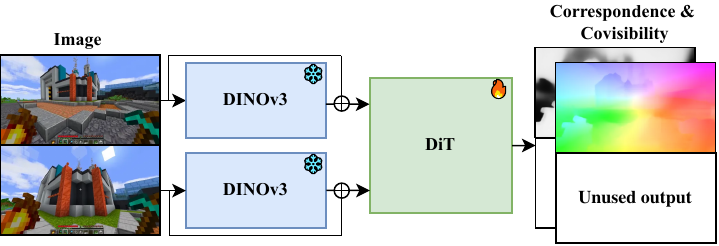}
    \caption{FreeMatching Architecture. Our transformer is initialized from FLUX.2-klein-base-4B. Each image is encoded into latent tokens and augmented with projected DINOv3 features. The two token sequences are concatenated for joint attention, and a task-specific decoder predicts the dense correspondence map and covisibility mask in a single forward pass.
    }
    \label{fig:model}
\end{figure}

To achieve this goal, we combine the complementary representations of two foundation models. We initialize our generative transformer with FLUX.2-klein-base-4B~\cite{bfl2026flux2klein} (FLUX for brevity) and inject DINOv3~\cite{simeoni2025dinov3} features to incorporate semantic information. The stage-wise results show improved performance when combining generative initialization with DINOv3 features.

As shown in~\cref{fig:model}, a frozen FLUX VAE encodes each image into latent tokens. We add linearly projected DINOv3 patch features to the corresponding token embeddings and concatenate the two image sequences. Position encodings retain spatial location and image identity, allowing joint attention to model cross-image relationships. During supervised pre-training, we freeze the foundation encoders and train the transformer, semantic projection, and task-specific output decoder. The model predicts the dense correspondence map $C$ and covisibility mask $M$ in a single forward pass. Architecture details are provided in~Appendix~\ref{app:architecture}.

\subsection{Supervised Pre-training}

\begin{table}[t]
    \centering
    \caption{\textbf{Supervised Training Data.} We combine classical correspondence datasets with our tracked-video and Blender data. Ratios indicate sampling proportions, rounded to multiples of 5\%.}
    \label{tb:pretrain_data}
    \fontsize{8pt}{10pt}\selectfont
    \renewcommand{\arraystretch}{1.12}
    \begin{tabular}{@{}p{0.25\linewidth}r@{\hspace{2.5em}}p{0.25\linewidth}r@{}}
    \toprule
    Dataset & Ratio & Dataset & Ratio \\
    \midrule
    ScanNet++~\cite{yeshwanth2023scannetpp} & 15\% & Spring~\cite{mehl2023spring} & 5\% \\
    BlendedMVS~\cite{yao2020blendedmvs} & 10\% & PointOdyssey~\cite{zheng2023pointodyssey} & 5\% \\
    FlyingThings~\cite{mayer2016large} & 10\% & FlyingChairs~\cite{fischer2015flownet} & 5\% \\
    DynamicReplica~\cite{karaev2023dynamicstereo} & 10\% & & \\
    \midrule
    Video (Ours) & 20\% & Blender (Ours) & 20\% \\
    \bottomrule
    \end{tabular}
\end{table}

\paragraph{Dataset Collection}

To establish a robust generalist foundation across a wide range of transformations, alongside standard optical flow and dense matching datasets, we introduce two new large-scale annotated datasets, capturing both real-world diversity and synthetic precision. Specifically, our new datasets comprise:

\begin{enumerate}
    \item Real-world Dynamic Scenes: We leverage a SOTA tracking model AllTracker~\cite{harley2025alltracker} to generate dense pseudo-ground truth for 520k clips from the ViPE~\cite{huang2025vipe} dataset. By sampling frame pairs with varying temporal intervals, we capture diverse non-rigid motion patterns that extend beyond the smooth assumptions of classical optical flow.
    \item Synthetic Objaverse Scenes: To mitigate inherent annotation noise, we render 360k image pairs with Blender using assets from Objaverse~\cite{deitke2023objaverse}. This provides pixel-perfect ground truth with semantic and structural diversity, serving as a high-fidelity anchor to complement the real-world data.
\end{enumerate}

The supervised data sources and their sampling proportions are summarized in~\cref{tb:pretrain_data}. Detailed descriptions of all datasets are provided in~Appendix~\ref{app:datasets}.

\paragraph{Robust Training Loss}

Since real-world datasets contain inevitable annotation noise, standard $\ell_2$ loss can destabilize training. We therefore adopt the robust Huber loss~\cite{huber1992robust} for correspondence $C$ regression, while maintaining $\ell_2$ for the covisibility masks $M$. Formally, we minimize:
\begin{equation}
    \begin{split}
        \mathcal{L}_{\text{pre}} = \frac{1}{N} \sum_{x,y} & \Bigl( M^*(x, y) \cdot \mathcal{L}_{\text{Huber}}(C(x, y), C^*(x, y)) + \lambda\|M(x, y) - M^*(x, y)\|_2^2 \Bigr)
    \end{split}
\end{equation}
where $C^*$ and $M^*$ denote the ground-truth correspondence and covisibility, respectively, and $N=H\times W$ is the number of pixels. The coefficient $\lambda$ balances mask supervision against correspondence regression; the loss weights are specified in Appendix~\ref{app:training}. This hybrid objective enables FreeMatching to learn precise matching from clean synthetic data while remaining robust to noisy real-world annotations. 

\subsection{Teacher-Guided Iterative Refinement}

\paragraph{Dataset Collection}

Supervised pre-training uses classical correspondence datasets, tracked videos, and synthetic scenes. To extend matching to IEG without dense correspondence annotations, we refine the model on image pairs from this domain. The experimental data and refinement configuration are specified in Appendix~\ref{app:training}.

\begin{figure}
    \centering
    \includegraphics[width=0.70\linewidth]{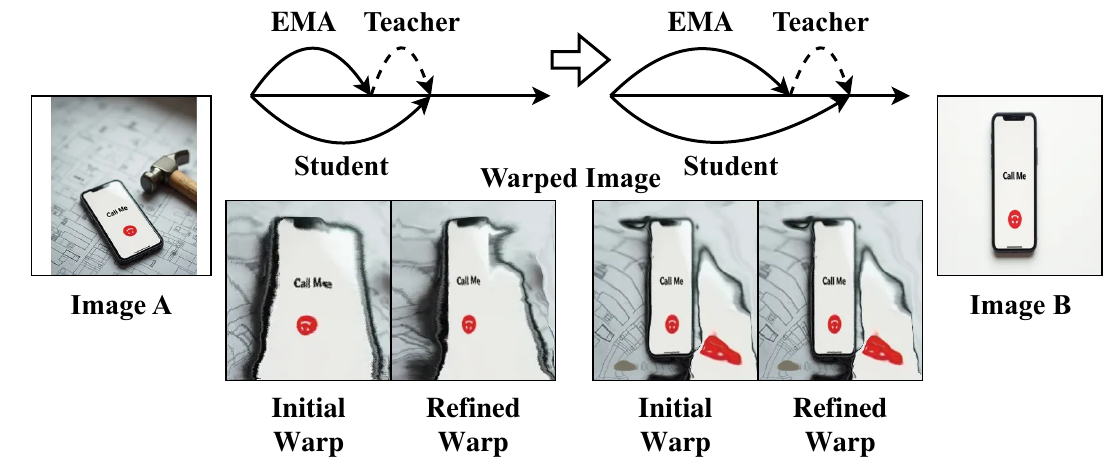}
    \caption{Teacher-Guided Iterative Refinement. An exponential moving average (EMA) copy of the student predicts the correspondence used to produce the Initial Warp. A frozen teacher estimates a local correction; composing it with the initial correspondence produces the Refined Warp and a correspondence target for training the student. The EMA copy evolves with the student while the teacher remains fixed.}
    \label{fig:weak-train}
\end{figure}

\paragraph{Refinement Procedure} 

To improve local alignment without dense correspondence annotations, we use a trainable student and a frozen refinement teacher. An exponential moving average (EMA) copy of the student produces the initial correspondence $C_0$, yielding a warped image $\hat{I}_{\mathrm{inter}}=I_a(C_0)$. The fixed teacher estimates a local correction map $S$ from the target $I_b$ to this intermediate image.

We compose the two maps to obtain the pseudo-target $C_{\mathrm{new}}=C_0\circ S$. The student learns to match this target in valid regions, while its covisibility prediction is regularized toward the EMA prediction. As the student improves, the EMA copy provides updated initial alignments for the fixed teacher to refine. This separates learning to bridge large changes from correcting the remaining local misalignment.

The refinement teacher can be instantiated with a pretrained correspondence model. Our concrete choice, EMA update, validity criteria, and losses are given in Appendix~\ref{app:training}.

\subsection{Consistency Evaluation}
\label{sec:method_metric}

Having established a robust generalist foundation via our two-stage training, we employ a reconstruction-based protocol that serves a dual purpose: evaluating correspondence quality and quantifying data consistency. The core premise is that an ideal correspondence map $C$ should allow the source image $I_a$ to be warped into $\hat{I}_b = I_a(C)$, aligning precisely with the target $I_b$ within the covisible region $M$. We quantify this alignment fidelity with separately reported reconstruction metrics:
\begin{equation}
    \text{Score}_{m} = \mathcal{D}_{m}(\hat{I}_b, I_b; M_m),
\end{equation}
where $m$ indexes the metric and $M_m$ specifies its evaluation region. We report MSE for pixel precision, LPIPS~\cite{zhang2018perceptual} for perceptual similarity, and feature similarities for semantic alignment, without combining them into a single score. Ref-MSE and Ref-DINOv2 use a shared frozen reference mask across methods; Full-MSE and whole-image OpenCLIP use the full image ($M_m\equiv1$). We assess the consistency of the results across pixel-level, perceptual, and feature-similarity metrics. DINOv3 scores and heatmaps quantify feature alignment, while MSE, LPIPS, DINOv2, SigLIP~2, and OpenCLIP provide complementary views of reconstruction quality. Definitions of all metrics, including SigLIP~2, are provided in~Appendix~\ref{app:evaluation}.

This enables a bi-directional evaluation paradigm:
\begin{enumerate}
    \item Evaluating Matching Accuracy: On benchmarks where image pairs are known to be consistent, the reconstruction error serves as a proxy for the quality of our predicted correspondence $C$.
    \item Evaluating Identity-Preserving Consistency: Conversely, assuming the matching model is sufficiently robust, this protocol can be inverted to evaluate the data itself. For new image pairs, significant reconstruction error within the covisible region can indicate inconsistencies in the image content, subject to the accuracy of the estimated alignment. Consequently, our model functions as a quantitative metric for identity preservation, enabling the objective assessment of consistency in novel data samples.
\end{enumerate}

\section{Experiments}
\label{sec:experiments}

In this section, we empirically validate FreeMatching. We first demonstrate its robustness by achieving competitive performance on classical correspondence benchmarks (optical flow and dense matching). We then showcase its distinct advantage on our proposed IEG-Bench, designed to evaluate correspondence between shared subjects in source or reference images and IEG outputs. Finally, we provide extensive ablation studies, qualitative visualizations, and an analysis of FreeMatching's utility as a perceptual metric.

For refinement, we instantiate the fixed teacher with RoMa~\cite{edstedt2024roma}. Experimental details are provided in Appendix~\ref{app:training}, with training and inference costs in Appendix~\ref{app:implementation}.

\begin{table*}[t!]
\centering
\caption{Dense Matching and Relative Pose Estimation. Comparison with representative correspondence methods on ETH3D~\cite{schops2017multi} and ScanNet-1500~\cite{dai2017scannet}. A single FreeMatching model achieves competitive performance across outdoor and indoor geometric matching and downstream pose estimation. Best and second-best results are \textbf{bold} and \underline{underlined}, respectively.}
\label{tab:dense}
\fontsize{7.5pt}{10pt}\selectfont
\setlength{\tabcolsep}{4pt}
\makebox[\textwidth][l]{\begin{tabular}{@{}l cccc cccc ccc ccc@{}}
\toprule
 & \multicolumn{8}{c}{Dense Matching} & \multicolumn{6}{c}{Pose Estimation (AUC, \%)} \\
\cmidrule(lr){2-9} \cmidrule(lr){10-15}
 & \multicolumn{4}{c}{ETH3D} & \multicolumn{4}{c}{ScanNet-1500} & \multicolumn{3}{c}{ETH3D} & \multicolumn{3}{c}{ScanNet-1500} \\
\cmidrule(lr){2-5} \cmidrule(lr){6-9} \cmidrule(lr){10-12} \cmidrule(lr){13-15}
Method & EPE$\downarrow$ & 1px$\downarrow$ & 2px$\downarrow$ & 5px$\downarrow$ & EPE$\downarrow$ & 1px$\downarrow$ & 2px$\downarrow$ & 5px$\downarrow$ & @5°$\uparrow$ & @10°$\uparrow$ & @20°$\uparrow$ & @5°$\uparrow$ & @10°$\uparrow$ & @20°$\uparrow$ \\
\midrule
GMFlow~\cite{xu2022gmflow} & 131.99 & 93.15 & 86.95 & 78.54 & 53.28 & 96.95 & 90.39 & 73.25 & 0.17 & 0.62 & 2.28 & 0.42 & 1.64 & 5.62 \\
SEA-RAFT~\cite{wang2024sea} & 150.79 & 90.65 & 86.21 & 80.93 & 67.83 & 95.80 & 88.03 & 71.53 & 0.35 & 1.25 & 3.83 & 1.40 & 3.94 & 9.62\\
UniMatch~\cite{xu2023unifying} & 112.97 & 82.20 & 75.09 & 67.71 & 44.64 & 94.57 & 84.05 & 60.83 & 0.80 & 2.79 & 7.26 & 1.91 & 5.12 & 12.02 \\
RoMa~\cite{edstedt2024roma} & \underline{11.26} & \underline{50.69} & 37.18 & 24.53 & \underline{12.03} & 91.91 & 76.48 & 43.31 & \underline{53.48} & \underline{68.32} & \textbf{78.61} & \textbf{31.80} & 53.40 & \textbf{70.90} \\
UFM~\cite{zhang2025ufm} & 12.98 & 54.01 & \underline{35.90} & \textbf{20.63} & \underline{12.03} & \underline{91.35} & \underline{74.57} & \textbf{37.52} & 38.77 & 53.59 & 66.05 & \underline{31.60} & \textbf{54.10} & - \\
\midrule
{FreeMatching (Ours)} & \textbf{10.23} & \textbf{50.53} & \textbf{35.67} & \underline{20.71} & \textbf{8.56} & \textbf{91.21} & \textbf{74.18} & \underline{38.66} & \textbf{54.22} & \textbf{68.76} & \underline{78.23} & 31.51 & \underline{53.62} & \underline{70.03} \\
\bottomrule
\end{tabular}}
\end{table*}

\begin{table*}[t!]
\centering
\caption{Optical Flow Estimation. Comparison with representative optical-flow and dense-matching methods on Sintel~\cite{butler2012naturalistic} and KITTI-2015~\cite{menze2015object}, evaluating the cross-task generalization of the same FreeMatching model. Best and second-best results are \textbf{bold} and \underline{underlined}, respectively.}
\label{tab:flow}
\fontsize{8pt}{10pt}\selectfont
\setlength{\tabcolsep}{5pt}
\begin{tabular}{@{}l cccc cccc cccc@{}}
\toprule
& \multicolumn{4}{c}{{Sintel (Clean)}} & \multicolumn{4}{c}{{Sintel (Final)}} & \multicolumn{4}{c}{{KITTI-2015}} \\
\cmidrule(lr){2-5} \cmidrule(lr){6-9} \cmidrule(lr){10-13}
Method & EPE$\downarrow$ & 1px$\downarrow$ & 2px$\downarrow$ & 5px$\downarrow$ & EPE$\downarrow$ & 1px$\downarrow$ & 2px$\downarrow$ & 5px$\downarrow$ & EPE$\downarrow$ & 1px$\downarrow$ & 2px$\downarrow$ & 5px$\downarrow$ \\
\midrule
GMFlow~\cite{xu2022gmflow} & 1.21 & 41.3 & 12.6 & 1.9 & 1.56 & 42.7 & 14.4 & 3.2 & 4.32 & 60.3 & 35.5 & 16.1 \\
SEA-RAFT~\cite{wang2024sea} & \underline{0.43} & \underline{5.9} & \underline{2.7} & \underline{0.9} & \underline{0.81} & \textbf{8.7} & \textbf{4.8} & \textbf{2.1} & \underline{1.97} & \underline{29.7} & \underline{16.8} & \underline{7.2} \\
UniMatch~\cite{xu2023unifying} & \textbf{0.40} & \textbf{5.4} & \textbf{2.4} & \textbf{0.8} & \textbf{0.75} & \underline{8.9} & \underline{4.9} & \underline{2.2} & \textbf{1.52} & \textbf{24.1} & \textbf{12.5} & \textbf{5.1} \\
RoMa~\cite{edstedt2024roma} & 1.24 & 25.2 & 8.1 & 4.5 & 2.49 & 35.6 & 16.6 & 8.7 & 5.91 & 41.4 & 27.5 & 18.4 \\
UFM~\cite{zhang2025ufm} & 0.92 & 14.2 & 6.4 & 2.1 & 1.48 & 16.9 & 9.0 & 3.8 & 4.75 & 39.6 & 24.5 & 15.5 \\
\midrule
{FreeMatching (Ours)} & 1.05 & 7.1 & 3.6 & 1.3 & 1.36 & 12.2 & 6.5 & 3.1 & 4.95 & 32.5 & 17.6 & 9.2\\
\bottomrule
\end{tabular}
\end{table*}

\subsection{Classical Benchmarks}
\label{sec:classical}

Tables~\ref{tab:dense} and~\ref{tab:flow} compare FreeMatching with representative classical correspondence methods. After 10k IEG refinement, FreeMatching achieves lower EPE than the similarly adapted UFM and RoMa variants on ScanNet and ETH3D. On optical flow, the adapted UFM is stronger on Sintel Clean and KITTI, while FreeMatching is stronger on Sintel Final; UniMatch achieves the lowest EPE among the methods shown. These results demonstrate how FreeMatching extends matching to IEG while maintaining competitive classical performance. The complete 10k adaptation comparison is reported in Table~\ref{tab:adapted_classical}. 

\subsection{Identity-Preserving Correspondence}
\label{sec:identity_preserving}

The next part, {IEG-Bench}, probes a model's core ability to establish correspondences while preserving an object's {identity}. To quantify this, we employ the reconstruction-based evaluation protocol described in~\cref{sec:method_metric} and Appendix~\ref{app:evaluation}. We compute a perceptual similarity score between the warped source and the target image. A higher score indicates a more accurate correspondence field that better preserves the object's identity.

As detailed in Table~\ref{tab:baseline_IEG}, our IEG-Bench challenges existing paradigms. While specialized models like SEA-RAFT and UniMatch are ill-equipped for this semantic task, contemporary generalists like RoMa and UFM demonstrate impressive generalization. Thanks to their more relaxed priors and powerful feature backbones, they achieve respectable performance, proving their robustness beyond classical tasks.

\begin{figure*}[t]
    \centering
    \includegraphics[width=\linewidth]{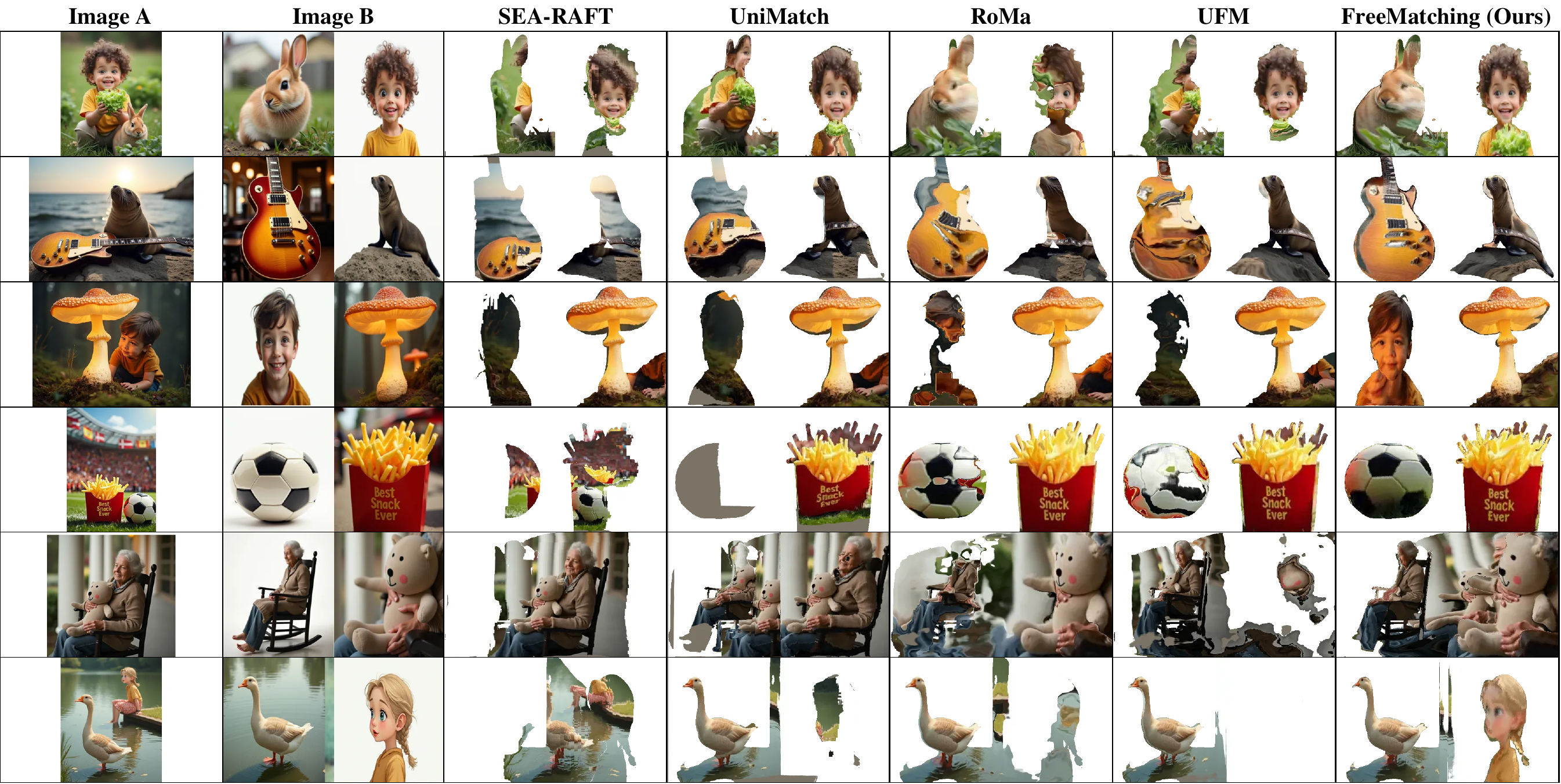}
    \caption{Visual comparisons on IEG-Bench using UNO examples. The references in Image B are composed into Image A ($B\rightarrow A$). Each method warps Image A toward Image B to evaluate correspondence quality.}
    \label{fig:baseline_visual}
\end{figure*}

RoMa and UFM nevertheless exhibit local distortions or incomplete matches in the IEG examples in \cref{fig:baseline_visual}. The adapted comparison below assesses whether these differences persist after the models receive IEG refinement data.

FreeMatching achieves the strongest reconstruction scores among the compared methods in Table~\ref{tab:baseline_IEG}. These gains support its ability to establish identity-preserving correspondences across challenging IEG image pairs. In Appendix~\ref{app:human_eval}, we use human evaluations to assess how our reconstruction-based metric correlates with human judgments of identity preservation.

\begin{table}[h]
\centering
\caption{IEG-Bench Evaluation. We report the mean reconstruction similarity scores (higher is better) using SigLIP~2~\cite{tschannen2025siglip} and DINOv3~\cite{simeoni2025dinov3} features, along with reference-mask MSE (Ref-MSE) and LPIPS~\cite{zhang2018perceptual} (lower is better) between the warped and target images. The identity baseline compares the unwarped source image with the target using an identity correspondence map. Best and second-best results are \textbf{bold} and \underline{underlined}, respectively.}
\label{tab:baseline_IEG}
\fontsize{8pt}{10pt}\selectfont
\setlength{\tabcolsep}{5pt}
\begin{tabular}{@{}lcccc@{}}
    \toprule
    {Model} & {Ref-MSE} $\downarrow$ & {LPIPS} $\downarrow$ & {SigLIP~2} $\uparrow$ & {DINOv3} $\uparrow$ \\
    \midrule
    Identity & 23.65 & 81.97 & 86.67 & 52.38 \\
    \midrule
    SEA-RAFT~\cite{wang2024sea} & 18.49 & 72.45 & 85.18 & 58.14 \\
    UniMatch~\cite{xu2023unifying} & 11.07 & 58.42 & \underline{87.86} & 66.35 \\
    RoMa~\cite{edstedt2024roma} & \underline{8.08} & \underline{48.91} & 87.73 & \underline{70.15} \\
    UFM~\cite{zhang2025ufm} & 10.41 & 53.05 & 86.97 & 66.77 \\
    \midrule
    {FreeMatching (Ours)} & \textbf{6.22} & \textbf{45.41} & \textbf{90.77} & \textbf{80.01} \\
    \bottomrule
\end{tabular}
\end{table}

\paragraph{Comparison after weak adaptation.}
We additionally compare FreeMatching, UFM, and RoMa after 10k adaptation updates using UNO+OmniGen2 pairs. Table~\ref{tab:adapted_ieg} reports reference-mask and full-image MSE, together with DINOv2 and OpenCLIP similarities as complementary feature metrics. FreeMatching achieves the best scores across all four metrics. Together with the MSE, LPIPS, SigLIP~2, and DINOv3 results in Table~\ref{tab:baseline_IEG}, these results show consistent advantages across metrics under both evaluation settings. Its paired Full-MSE advantages over UFM and RoMa are 2.48 (95\% CI: [1.54, 3.43]) and 5.57 ([4.06, 7.13]), respectively. Evaluation definitions and paired statistics are given in Appendix~\ref{app:independent_metrics}.

\begin{table}[t]
\centering
\caption{IEG-Bench comparison after 10k adaptation updates. All variants use UNO+OmniGen2 refinement data. Ref-MSE and Ref-DINOv2 use the shared frozen reference mask; Full-MSE and OpenCLIP use the full image. All metrics are scaled by 100. DINOv2~\cite{oquab2023dinov2} and OpenCLIP~\cite{cherti2023scaling} complement the DINOv3 metric used in the stage-wise analysis.}
\label{tab:adapted_ieg}
\fontsize{8pt}{10pt}\selectfont
\setlength{\tabcolsep}{4pt}
\begin{tabular}{@{}lcccc@{}}
\toprule
Method & Ref-MSE $\downarrow$ & Full-MSE $\downarrow$ & Ref-DINOv2 $\uparrow$ & OpenCLIP $\uparrow$ \\
\midrule
FreeMatching (10k-IEG) & \textbf{6.22} & \textbf{10.99} & \textbf{54.31} & \textbf{70.56} \\
UFM (10k-IEG)~\cite{zhang2025ufm} & 8.55 & 13.48 & 39.65 & 56.48 \\
RoMa (10k-IEG)~\cite{edstedt2024roma} & 9.70 & 16.57 & 47.14 & 61.18 \\
\bottomrule
\end{tabular}
\end{table}

\subsection{Qualitative Results}
\label{sec:qualitative}

Figure~\ref{fig:baseline_visual} illustrates the matching challenges behind the quantitative results. The examples require preserving instance-level correspondences across substantial changes in appearance, pose, and composition. The compared methods exhibit local distortions or incomplete matches, while FreeMatching produces more coherent alignments in these examples.

\subsection{Fine-Grained Identity Assessment}
\label{sec:consistency}

Leveraging FreeMatching's precise alignment, we extend its utility to act as a fine-grained metric for identity preservation. Following \cref{sec:method_metric}, by warping the source image to the target and computing local similarity strictly within the covisible region, we generate a spatial heatmap of perceptual divergence.

This approach addresses a critical blind spot of global metrics~\cite{huang2024vbench} (e.g., global DINO or CLIP scores). While global metrics effectively measure high-level semantic coherence, they inherently aggregate spatial information, lacking the capability to provide granular feedback on local details. Consequently, they often mask subtle yet critical deviations in texture or structure that FreeMatching reveals.

\begin{figure*}[t]
    \vspace{-10pt}
    \centering
    \includegraphics[width=\linewidth]{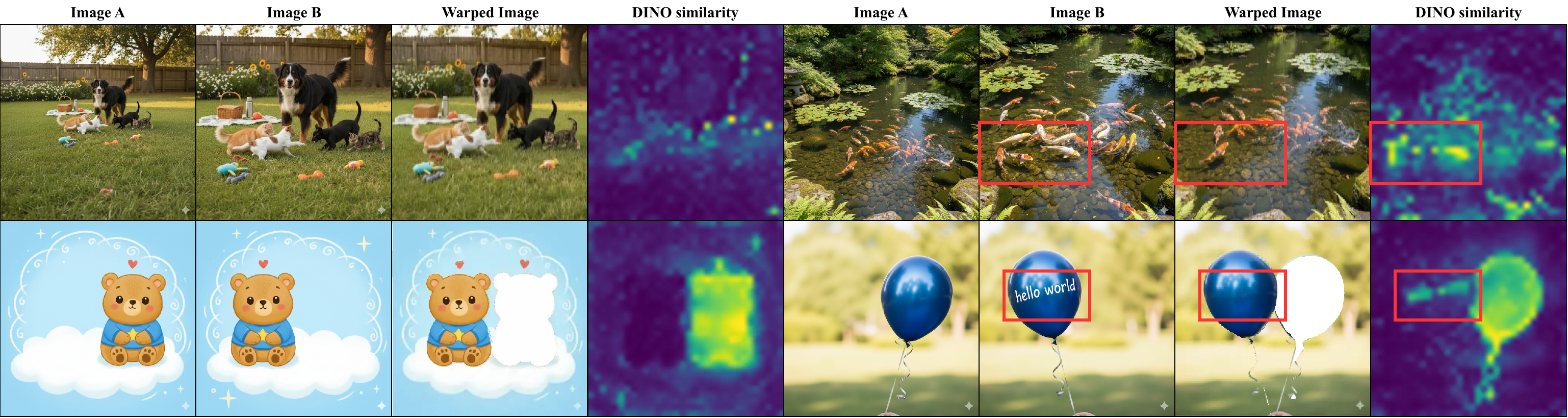}
    \caption{Fine-Grained Identity Assessment. Nano Banana edits Image A into Image B ($A\rightarrow B$). DINOv3 similarity heatmaps compare the warped source with the target within covisible regions, revealing local appearance inconsistencies. Red boxes highlight affected regions, including the altered balloon text.}
    \label{fig:application}
\end{figure*}

\cref{fig:application} illustrates this spatially resolved assessment. In the second row, the heatmap highlights the inconsistency in the balloon's text. By showing where local appearance departs from the reference after alignment, FreeMatching provides spatial feedback for assessing identity preservation in generative editing.

\subsection{Stage-wise System Construction}
\label{sec:ablation}

\begin{table}[t]
    \centering
    \caption{{Stage-wise System Construction.} We progressively add components and report DINOv3 similarity ($\times 100$) on IEG-Bench. The checkmark (\checkmark) indicates the component is included. {FLUX Init}: FLUX generative initialization; {DINOv3}: DINOv3 semantic features; {Video \& Blender}: Video \& Blender datasets; {Refine}: Teacher-guided iterative refinement.}
    \label{tab:ablation}
    \fontsize{8pt}{10pt}\selectfont
    \setlength{\tabcolsep}{3.5pt} \begin{tabular}{c|cccc|c}
    \toprule
    {ID} & {FLUX Init} & {DINOv3} & {Video \& Blender} & {Refine} & {DINOv3} $\uparrow$ \\
    \midrule
    1 &            &  &  &  & 67.31 \\
    2 & \checkmark &            &  &  & 69.28 \\
    3 & \checkmark & \checkmark &            &  & 72.76 \\
    4 & \checkmark & \checkmark & \checkmark &            & 76.52 \\
    \midrule
    {5} & {\checkmark} & {\checkmark} & {\checkmark} & {\checkmark} & \textbf{80.01} \\
    \bottomrule
    \end{tabular}
\end{table}

Table~\ref{tab:ablation} reports the cumulative construction of FreeMatching. ID1 is a randomly initialized base DiT trained on classical supervised datasets with Huber correspondence loss and an $\ell_2$ mask loss; it uses neither foundation-model initialization nor DINOv3 inputs, Video/Blender data, or refinement. Subsequent rows add FLUX initialization, DINOv3 features, Video/Blender supervision, and teacher-guided iterative refinement in sequence. DINOv3 reconstruction similarity increases across these stages, measuring the cumulative improvement in feature alignment. The results show that:
\begin{enumerate}
    \item FLUX initialization and DINOv3 features each improve matching in the cumulative construction, supporting the use of generative and semantic foundation representations.
    \item Adding our video and Blender supervision further improves matching on IEG-Bench, extending the capabilities learned from classical correspondence datasets.
    \item Teacher-guided iterative refinement further improves matching on IEG, complementing the gains from foundation-model representations and heterogeneous supervision.
\end{enumerate}
\paragraph{Refinement across domains.}
Table~\ref{tab:refinement_domains} compares Stage~1 with refinement on unlabeled flow-domain pairs (10k-Flow) and IEG pairs (10k-IEG). Both improve IEG-Bench Ref-MSE over Stage~1. The gains depend on the refinement domain: 10k-Flow performs best on Sintel Clean, while 10k-IEG performs best on the remaining reported benchmarks. Compared with Stage~1, 10k-IEG also lowers EPE on all five classical benchmarks, showing that the extension to IEG retains classical matching capabilities.

\section{Discussion}
In this work, we introduced FreeMatching, a generalizable framework that extends dense correspondence matching beyond classical geometric and motion settings to IEG. By combining foundation-model representations, heterogeneous correspondence supervision, and teacher-guided iterative refinement, a single model establishes identity-preserving correspondences across diverse visual transformations while retaining competitive performance on classical benchmarks. This broader capability also enables spatially resolved assessment of identity preservation in IEG. We envision a mutually beneficial future where generative models provide diverse data for correspondence learning, while robust correspondence serves as a metric for guiding and evaluating image editing and reference-guided generation.

\paragraph{Generative Ambiguity}
Generative tasks can also introduce ambiguous correspondences, for example through object replication. In~Appendix~\ref{app:one_to_many}, we show how our framework can be extended to a probabilistic diffusion formulation for modeling such ambiguities. This formulation provides a route to representing multiple plausible matches, while the experiments in this paper use the efficient single-pass regression model.

\paragraph{Semantic Matching}
Semantic correspondence methods such as SD-DINO~\cite{zhang2023tale} align semantically related image regions, including across different instances of a category. Our objective instead emphasizes preserving the identity and local details of the same instance across IEG image pairs. The qualitative comparison in~Appendix~\ref{app:semantic_matching} illustrates how FreeMatching preserves object shape and appearance more consistently in the displayed multi-object example.

\paragraph{Limitations}
Despite these advances, limitations remain. First, the model can be challenged by extreme cases, such as severe occlusions, topological changes, or drastic non-rigid deformations. Second, its performance ceiling is intrinsically tied to the capabilities of the underlying pre-trained foundation models. Finally, our IEG-Bench, while a valuable starting point, would benefit from expansion to cover more diverse categories. Addressing these challenges presents exciting avenues for future research.

\bibliographystyle{unsrtnat}
\bibliography{references}

\newpage
\appendix
\section*{Appendix}

\section{Architecture}
\label{app:architecture}

In this section, we provide the detailed architectural specifications. Our framework leverages the generative prior of a Diffusion Transformer (DiT) and the semantic robustness of DINOv3 to predict dense correspondence and covisibility masks.

\subsection{Architecture Overview}

\paragraph{Backbone.} We initialize the generative transformer from \texttt{FLUX.2-klein-base-4B}~\cite{bfl2026flux2klein}, an undistilled rectified-flow model. We retain its latent-token input projection and add semantic feature injection and a task-specific output decoder for correspondence prediction.

\paragraph{Image Compression (VAE).} The frozen FLUX VAE deterministically encodes each image $I\in\mathbb{R}^{H\times W\times3}$ into a latent representation $z\in\mathbb{R}^{h\times w\times32}$, where $h=H/8$ and $w=W/8$. We pack each $2\times2$ latent neighborhood into a 128-dimensional token and normalize it using the VAE's stored statistics. At input resolution $H\times W=288\times512$, each image yields a $36\times64$ latent grid and $18\times32=576$ packed tokens.

\paragraph{Semantic Feature Extraction.} A frozen DINOv3 ViT-L/16 encoder extracts 1024-dimensional patch features at the same $H/16\times W/16$ spatial grid. We discard the class and register tokens, then use a zero-initialized linear layer to project the patch features to the transformer's hidden dimension. These features are added to the corresponding projected latent tokens.

\subsection{Input Representation and Conditioning}

The source and target token sequences are concatenated along the sequence dimension, yielding 1152 image tokens at our input resolution. FLUX position encodings retain both spatial coordinates and image identity, enabling attention across the two views. We use cached empty-prompt embeddings for the text-conditioning path; correspondence prediction requires only the image pair.

\subsection{Prediction and Output Decoding}

The transformer predicts output tokens in a single forward pass. We select the token sequence associated with the correspondence query image, reverse the latent normalization, and unpack the tokens into a 32-channel latent grid. A task-specific convolutional decoder with progressive upsampling produces a full-resolution three-channel output: two channels for correspondence coordinates and one for covisibility. The predicted coordinates are converted to pixel coordinates to obtain $C\in\mathbb{R}^{H\times W\times2}$, with the remaining channel providing $M\in\mathbb{R}^{H\times W\times1}$. The parameter-update scope for both training stages is specified in Appendix~\ref{app:implementation}.

Table~\ref{tab:inference_cost} reports measured inference latency and peak allocated memory under a common external input resolution on an H20 GPU. Improving inference efficiency remains a direction for future work.

\section{Datasets}
\label{app:datasets}

\begin{figure}[t]
\centering
\includegraphics[width=0.40\linewidth]{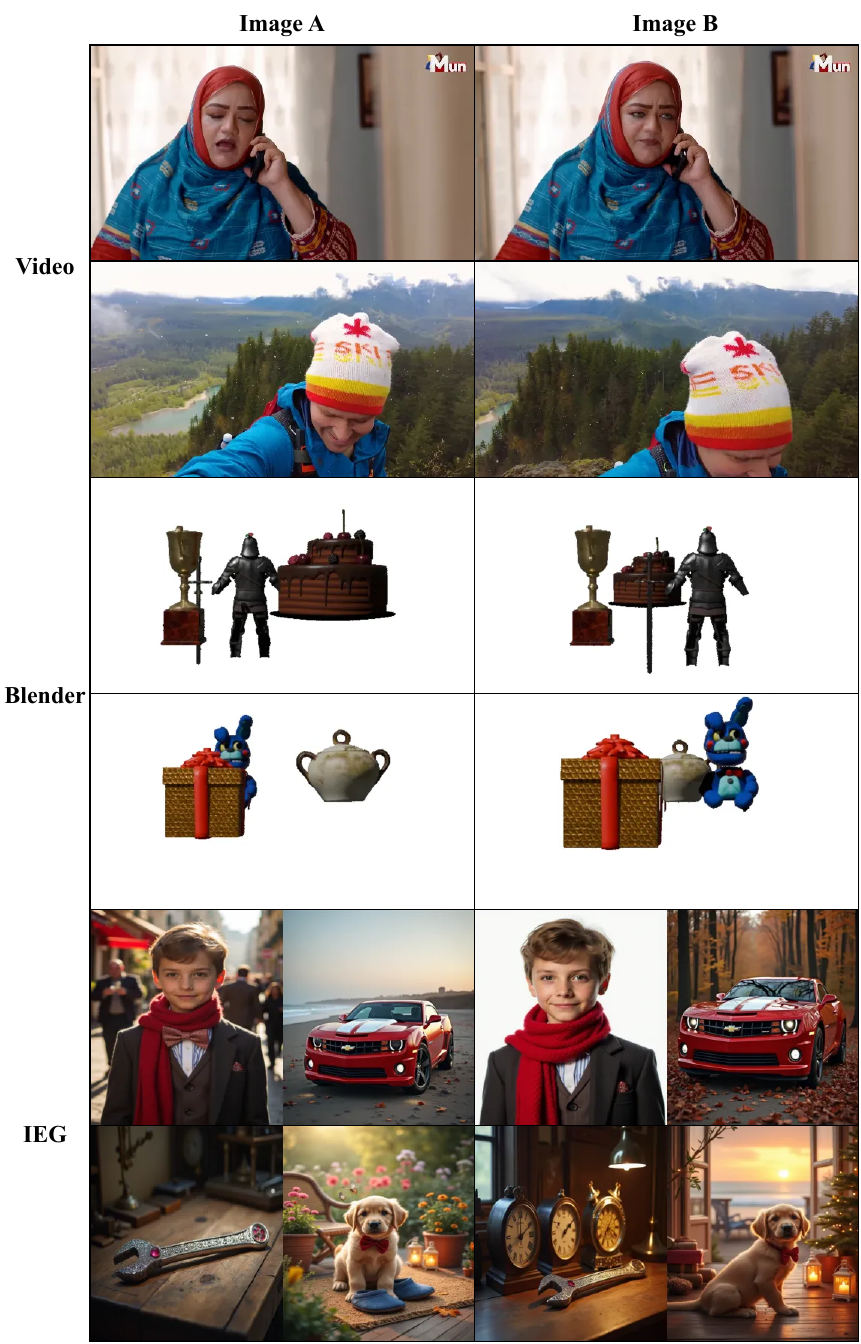}
\caption{{Overview of the constructed datasets.} We utilize {Real-World Video} data for learning natural deformations, {Synthetic Blender} data for precise geometric supervision, and the {IEG} benchmark for evaluating identity-preserving correspondence between source or reference images and generated outputs.}
\label{fig:dataset}
\end{figure}

We present a comprehensive overview of the data strategy employed for pre-training and evaluation, combining established benchmarks with novel large-scale datasets to ensure both robustness and generalization. Some visual examples are provided in~\cref{fig:dataset}.

\subsection{Standard Datasets}
Our supervised mixture includes BlendedMVS~\cite{yao2020blendedmvs} and ScanNet++~\cite{yeshwanth2023scannetpp} for geometric correspondence, FlyingThings~\cite{mayer2016large}, Spring~\cite{mehl2023spring}, and FlyingChairs~\cite{fischer2015flownet} for optical flow, and PointOdyssey~\cite{zheng2023pointodyssey} and DynamicReplica~\cite{karaev2023dynamicstereo} for synthetic video correspondence. We follow the relevant preprocessing protocols in UFM~\cite{zhang2025ufm}. These sources complement our tracked-video and Blender data. Following Table~\ref{tb:pretrain_data}, the approximate sampling proportions are 10\% BlendedMVS, 15\% ScanNet++, 10\% FlyingThings, 5\% Spring, 5\% PointOdyssey, 10\% DynamicReplica, 5\% FlyingChairs, 20\% Video, and 20\% Blender. These proportions describe sampling frequency rather than dataset size and are rounded to 5-percentage-point increments.

\subsection{Real-World Video Dataset}
To capture real-world non-rigid deformations, we annotate 520k clips from the ViPE~\cite{huang2025vipe} dataset. We generate dense pseudo-ground truth annotations using the AllTracker~\cite{harley2025alltracker} model. By randomly sampling frame pairs from video clips, our training data encompasses diverse motion patterns and natural occlusions that static datasets typically lack. Due to the data volume and copyright restrictions on the source videos, we do not distribute these videos or their precomputed trajectory annotations. We provide data-loading and trajectory-format conversion code, together with preparation instructions for videos that users obtain and are authorized to use.

Specifically, we utilize the official AllTracker checkpoints from \url{https://github.com/aharley/alltracker}. To handle long-range dependencies, we employ a sliding-window inference mode with a window length of $L=16$ frames. The model performs $N_{\text{iters}}=4$ iterative updates per step to refine flow estimates, ensuring precise motion propagation. It outputs dense 2D trajectories $\mathbf{P} \in \mathbb{R}^{T \times H \times W \times 2}$ and visibility confidence maps $\mathbf{V} \in \mathbb{R}^{T \times H \times W}$. We utilize these visibility masks to handle occlusions, ensuring that the loss is computed only on valid pixel correspondences.

\subsection{Synthetic Blender Dataset}
To mitigate the noise inherent in pseudo-labeled real-world data, we construct a synthetic dataset of 360k image pairs using 3D assets from Objaverse~\cite{deitke2023objaverse}. Each scene is composed of randomly selected objects (2--3 per scene) subject to random rigid transformations. By rendering these scenes from varied viewpoints, we obtain precise pixel-level correspondence and covisibility masks. This provides a clean signal for learning occlusion handling without ambiguity.

\paragraph{Data Curation and Preprocessing.} To ensure high-quality geometric supervision, we implement a rigorous automated filtering pipeline for the raw Objaverse assets. First, we normalize the scale of all objects such that their bounding box diagonal equals a fixed unit length ($d=2.0$). To filter out assets with poor geometry (e.g., flat planes or needle-like structures), we enforce an aspect ratio threshold of $3.0$ on the bounding box dimensions; objects exceeding this ratio are either discarded or anisotropically scaled to meet the criteria. Furthermore, we compute the ratio between the convex hull area of the object's projected silhouette and its 2D bounding box area. Assets with a ratio below $0.2$ are rejected to ensure substantial pixel occupancy and avoid sparse wireframe-like structures.

\paragraph{Dynamic Scene Composition.} We simulate challenging dynamic scenarios involving independent rigid body motions. Each scene is initialized with $N$ objects ($N \in \{2, 3\}$) arranged in a randomized layout (linear for pairs, triangular for triplets). The motion generation involves two temporal frames:
\begin{enumerate}
    \item {Initialization (Frame 1):} Objects are placed with random spatial jitter and subjected to random 3D rotations (max tilt $30^{\circ}$) and scaling factors $\sigma \in [0.8, 1.2]$.
    \item {Motion Modeling (Frame 2):} To simulate large displacements, we apply independent rigid transformations to each object relative to Frame 1. This includes a rotation delta magnitude sampled uniformly from $[30^{\circ}, 50^{\circ}]$ and further scale variation ($0.8\times$ to $1.2\times$) to simulate depth changes.
\end{enumerate}

\paragraph{Adaptive Rendering Pipeline.} Unlike static camera setups, we employ an adaptive camera control system. For each scene, the camera tracks the centroid of the collective bounding box of all objects, dynamically adjusting its distance to ensure all targets remain within the viewing frustum. Rendering is performed using the Blender Cycles engine with 128 samples per pixel at a resolution of $640 \times 360$. We leverage Blender's compositor nodes to export ground-truth data: optical flow is extracted directly from the motion vector pass and saved as high-precision 32-bit OpenEXR files, while occlusion masks are generated from the alpha channel, providing ambiguity-free supervision for motion learning.

\subsection{IEG Dataset}
We curate the image editing and reference-guided generation (IEG) dataset based on UNO~\cite{wu2025less} to further extend the scope of dense correspondence matching.

For UNO data preparation, we annotate shared object regions through semantic concept extraction and mask generation. These object-region annotations are distinct from the correspondence-validity masks used by the EMA-based refinement objective described in Appendix~\ref{app:training}.

\paragraph{0. Multi-Object Composition.}
To enhance semantic complexity and prevent overfitting to center-bias, we modify the standard UNO training format. Specifically, we concatenate two distinct reference--generated image pairs into a single composite training instance. This forces the model to learn correspondence across multiple objects and cluttered semantics rather than simple single-object alignment.

\paragraph{1. Shared Concept Extraction.}
Given a reference--generated image pair $(I_a, I_b)$, we first identify the primary object categories shared between the two images. We employ {Qwen3-VL-8B}~\cite{bai2025qwen3vl} to perform Open-ended Visual Question Answering. We prompt the model with both images and the instruction: {``Identify the 1--2 most prominent common objects present in both images and output their class names.''} This yields a set of text labels $\mathcal{L} = \{l_1, \dots, l_k\}$ (where $k \in \{1, 2\}$).

\paragraph{2. Union Mask Generation.}
Using the extracted labels $\mathcal{L}$ as text prompts, we leverage {SAM~3}~\cite{carion2025sam} to segment the corresponding regions in both $I_a$ and $I_b$. Since a single label (e.g., ``car'') may correspond to multiple instance masks in the scene, we perform a pixel-wise binary union to aggregate all instances associated with $\mathcal{L}$. This results in unified binary masks $M_a$ and $M_b$, where $M(\mathbf{p}) = 1$ indicates the presence of the target objects and $0$ denotes the background.

\section{Experimental Setup}
\label{app:training}

\subsection{Benchmarks and Baselines}

\paragraph{Benchmarks} Our evaluation spans two major domains of correspondence matching.
\begin{itemize}
    \item \textbf{Classical Correspondence:} We evaluate on a comprehensive set of datasets to ensure generalization. For dense matching, we use the outdoor {ETH3D}~\cite{schops2017multi} and indoor {ScanNet-1500}~\cite{dai2017scannet}. For optical flow, we use the synthetic {Sintel}~\cite{butler2012naturalistic} and the real-world {KITTI-2015}~\cite{menze2015object}. These benchmarks are governed by strong geometric and motion priors. \\
    For most benchmarks, we report the End-Point Error ({EPE}) and the percentage of pixels with errors greater than certain thresholds ({1px, 2px, 5px}). We compute these metrics only on covisible regions and recompute the results ourselves.
    For relative pose estimation, we follow RoMa's pipeline and report pose-error AUC at $5^{\circ}$, $10^{\circ}$, and $20^{\circ}$. Pose error is the maximum of the angular errors in rotation and translation direction. AUC is the area under the cumulative pose-error curve up to each threshold, normalized by that threshold and multiplied by 100; higher is better. Unreported baseline entries are left blank.
    \item \textbf{Identity-Preserving Correspondence:} For this domain, we leverage the UNO~\cite{wu2025less} dataset, which features two {disjoint} subsets, ensuring a strict separation between training and evaluation. We utilize the {single-reference} subset exclusively for our teacher-guided iterative refinement stage. Conversely, our test set, the \textbf{IEG Correspondence Benchmark (IEG-Bench)}, is constructed from approximately 100 challenging image pairs drawn from the held-out {multi-reference} subset. To evaluate performance, we follow the approach in \cref{sec:method_metric}, using reconstruction consistency to assess correspondence quality.
\end{itemize}

\paragraph{Baselines} We compare against baselines across the correspondence spectrum.
\begin{itemize}
    \item {For classical correspondence,} we benchmark against top performers like {GMFlow}~\cite{xu2022gmflow}, {SEA-RAFT}~\cite{wang2024sea}, {UniMatch}~\cite{xu2023unifying}, RoMa, and UFM~\cite{zhang2025ufm}.  \item For IEG correspondence, we compare SEA-RAFT, UniMatch, RoMa, and UFM on IEG-Bench. We additionally report a 10k IEG adaptation comparison for FreeMatching, UFM, and RoMa in Table~\ref{tab:adapted_ieg}.
\end{itemize}

Our training pipeline consists of supervised pre-training on classical datasets, tracked videos, and synthetic scenes, followed by teacher-guided iterative refinement on IEG pairs.

\subsection{Supervised Pre-training}

In the first stage, we train the model using a composite loss function that enforces geometric consistency and mask accuracy on the heterogeneous correspondence datasets. Since real-world applications often involve noise, standard $\ell_2$ loss can destabilize training. We therefore adopt a robust formulation. The total supervised loss $\mathcal{L}_{\text{sup}}$ is defined as:
\begin{equation}
    \mathcal{L}_{\text{sup}} = \lambda_{\text{huber}} \mathcal{L}_{\text{huber}} + \lambda_{\text{mask}} \mathcal{L}_{\text{mask}}
\end{equation}
where we set $\lambda_{\text{huber}}=10$ and $\lambda_{\text{mask}}=0.2$. The implementation uses the coordinate encoding and per-image normalization detailed below.

\paragraph{Robust Correspondence Loss ($\mathcal{L}_{\text{huber}}$)}
For training, the absolute source coordinates $C=(u,v)$ are encoded as $\widetilde C=(2u/(W-1)-1,\,2v/(H-1)-1)$. Let $V$ denote the ground-truth-derived mask used for correspondence supervision. For each image pair, the supervised pre-training loss sums the two coordinate components and normalizes by the number of valid pixels:
\begin{equation}
    \mathcal{L}_{\text{huber}} = \frac{\sum_{x,y} V(x,y)\sum_{d=1}^{2}\mathcal{L}_{\text{H}}\bigl(\widetilde C_{\text{pred},d}(x,y),\widetilde C_{\text{gt},d}(x,y);\delta\bigr)}{\sum_{x,y} V(x,y)+\epsilon}.
\end{equation}
The batch loss averages these per-pair values. During Stage~2, correspondence losses additionally average the two coordinate components, giving a denominator of $2\sum_{x,y}V(x,y)+\epsilon$ for the component-wise sum.

We adapt the Huber threshold using an exponential moving average $e$ of the masked L1 error, with the same pixel and coordinate normalization as the correspondence loss. We initialize $e_0=0.05$ and update $e_t=0.998e_{t-1}+0.002\operatorname{MAE}_t$, then set $\delta_t=2e_t$. Stage~2 maintains separate error averages for supervised and weakly supervised updates.

\paragraph{Mask Loss ($\mathcal{L}_{\text{mask}}$)}
The task-level covisibility mask is defined in $[0,1]$, while its training target is encoded as $\widetilde M^*=2M^*-1$. With raw mask prediction $\widetilde M_{\text{pred}}$, the implementation computes
\begin{equation}
    \mathcal{L}_{\text{mask}} = \frac{1}{HW} \sum_{x,y} A(x,y)\bigl\|\widetilde M_{\text{pred}}(x,y)-\widetilde M^*(x,y)\bigr\|_2^2,
\end{equation}
where $A$ selects available mask supervision. Covisibility thresholds are applied after converting the raw prediction back to the $[0,1]$ convention via $(\widetilde M_{\text{pred}}+1)/2$.

\subsection{Teacher-Guided Iterative Refinement}

We initialize the student from the supervised checkpoint and refine it using UNO+OmniGen2 image pairs without dense correspondence annotations.

\paragraph{EMA initialization and fixed refinement teacher.}
An EMA copy of FreeMatching predicts the initial correspondence $C_0$ and covisibility $\bar M$. After each student update, its parameters are updated as $\bar\theta\leftarrow\beta\bar\theta+(1-\beta)\theta$, with $\beta=0.95$. We instantiate the frozen refinement teacher with RoMa~\cite{edstedt2024roma}. RoMa estimates a correction map $S$ between the EMA-warped source and the target, producing $C_{\mathrm{new}}=C_0\circ S$. The teacher weights remain fixed, and gradients do not propagate through pseudo-target generation.

\paragraph{Refinement objective.}
For weakly supervised image pairs, we minimize
\begin{equation}
    \mathcal{L}_{\mathrm{weak}}=10\mathcal{L}_{\mathrm{self}}+0.2\mathcal{L}_{\mathrm{mask\mbox{-}EMA}}.
\end{equation}
The validity mask $V$ is the intersection of the RoMa confidence mask (threshold 0.2) and the EMA covisibility mask (threshold 0.7). The correspondence loss is
\begin{equation}
    \mathcal{L}_{\mathrm{self}} = \frac{\sum_{x,y}V(x,y)\,\mathcal{L}_{\mathrm{Huber}}\bigl(\widetilde C_\theta(x,y),\operatorname{sg}[\widetilde C_{\mathrm{new}}(x,y)]\bigr)}{\sum_{x,y}V(x,y)+\epsilon},
\end{equation}
where $\widetilde C$ denotes coordinates normalized to $[-1,1]$, the Huber loss averages over the two coordinate components, and $\operatorname{sg}$ denotes stop-gradient. The mask-consistency term is
\begin{equation}
    \mathcal{L}_{\mathrm{mask\mbox{-}EMA}} = \frac{1}{HW}\sum_{x,y}\left\|\widetilde M_\theta(x,y)-\operatorname{sg}[\widetilde{\bar M}(x,y)]\right\|_2^2,
\end{equation}
with $\widetilde M=2M-1$. This term regularizes the student's covisibility prediction toward the EMA output. A batch with no valid pseudo-correspondences contributes no refinement loss. The reported refinement objective uses these correspondence and mask targets, without an additional object-containment loss.

\subsection{Implementation Details}
\label{app:implementation}

\paragraph{Parameter updates.}
During the selected FreeMatching refinement run, the transformer and semantic projection are updated, while the output decoder is frozen. The foundation feature extractors and RoMa refinement teacher also remain frozen; the EMA copy is updated only by exponential averaging.

The selected FreeMatching checkpoint is the EMA model after 10k total Stage-2 updates, initialized from the supervised checkpoint at step 275k. Refinement combines supervised training with weakly supervised updates on UNO+OmniGen2 pairs. This run uses BF16 mixed precision and a global batch size of 32, with the following settings:
\begin{itemize}
    \item \textbf{Optimizer:} AdamW with $\beta_1=0.9, \beta_2=0.999$, and weight decay $1e^{-2}$.
    \item \textbf{Learning Rate:} $3\times10^{-5}$, with 3000 warmup steps followed by a constant learning rate.
    \item \textbf{Timestep Conditioning:} We use a fixed timestep $t=620$ for single-pass correspondence prediction.
    \item \textbf{Memory:} The training supports gradient checkpointing to reduce GPU memory usage.
\end{itemize}

\paragraph{Original training cost.}
The original training configuration reported in the rebuttal uses 150k Stage-1 updates followed by 10k Stage-2 updates on 16 A100-40GB GPUs. Stage~1 takes approximately 2 days (768 A100 GPU-hours), and Stage~2 takes approximately 5 hours (80 A100 GPU-hours). These costs describe that training configuration, rather than measurements of the selected 275k-initialized refinement run.

\paragraph{Baseline adaptation cost.}
The 10k IEG adaptation experiments use 16 H20 GPUs and a global batch size of 32, with external image resolution $288\times512$. Table~\ref{tab:adaptation_cost} reports their approximate training costs. These runs correspond to the additional adaptation comparison in Table~\ref{tab:adapted_ieg} and are distinct from the original A100 training above.

\begin{table}[ht]
\centering
\small
\caption{Training cost of 10k IEG adaptation on 16 H20 GPUs with global batch size 32. GPU-hours are approximate totals across the training GPUs.}
\label{tab:adaptation_cost}
\begin{tabular}{@{}lc@{}}
\toprule
Method & Training cost (H20 GPU-hours) \\
\midrule
FreeMatching & 42.0 \\
UFM~\cite{zhang2025ufm} & 16.8 \\
RoMa~\cite{edstedt2024roma} & 30.8 \\
\bottomrule
\end{tabular}
\end{table}

\paragraph{Inference profiling.}
Table~\ref{tab:inference_cost} reports a separate inference profile on a single H20 GPU at external input resolution $288\times512$ and batch size 1. We use 10 warm-up calls followed by 50 timed calls to each model's inference wrapper, with CUDA synchronization around the timed loop. We report mean latency per image pair and peak allocated GPU memory.

\begin{table}[ht]
\centering
\small
\caption{Inference cost on one H20 GPU at external resolution $288\times512$ and batch size 1. Latency is averaged over 50 calls after warm-up; memory is peak allocated GPU memory.}
\label{tab:inference_cost}
\begin{tabular}{@{}lcc@{}}
\toprule
Method & Latency (ms/pair) & Peak memory (GiB) \\
\midrule
FreeMatching & 121.9 & 15.18 \\
UFM~\cite{zhang2025ufm} & 51.3 & 2.35 \\
RoMa~\cite{edstedt2024roma} & 102.5 & 2.58 \\
\bottomrule
\end{tabular}
\end{table}

\section{Objective Evaluation}
\label{app:evaluation}

In this section, we provide a detailed formulation of the reconstruction-based evaluation protocol used in our experiments. As outlined in the main text, our goal is to quantify the fidelity of the warped source image $\hat{I}_b = \mathcal{W}(I_a, C)$ against the ground-truth target image $I_b$, restricted to the covisible region defined by the binary mask $M$.

To ensure a holistic assessment covering pixel-level accuracy, perceptual quality, and semantic identity preservation, we report four complementary metrics separately, supplemented by the reference-mask, full-image, and alternative-feature evaluations in Appendix~\ref{app:independent_metrics}.

\subsection{Pixel-Level Fidelity: MSE}
To measure low-level structural alignment, we compute the Mean Squared Error (MSE) within the masked region. Let $\Omega = \{p \mid M(p) = 1\}$ denote the set of valid pixels in the covisible area. The MSE is defined as:
\begin{equation}
    \text{MSE}(\hat{I}_b, I_b, M) = \frac{1}{3|\Omega|} \sum_{p \in \Omega} \| \hat{I}_b(p) - I_b(p) \|_2^2,
\end{equation}
where pixel values are normalized to the range $[0, 1]$ and the factor of 3 averages over RGB channels. In our reported tables, we scale this value by $100$ for readability. Lower MSE indicates tighter spatial alignment of edges and textures.

\subsection{Perceptual Similarity: LPIPS}
Pixel-wise metrics like MSE are often overly sensitive to slight misalignments or high-frequency noise that do not affect human perception. To capture perceptual similarity, we utilize the Learned Perceptual Image Patch Similarity (LPIPS) metric~\cite{zhang2018perceptual}.

We employ the AlexNet-based backbone. Unlike the standard global LPIPS, we compute the \textbf{spatial} LPIPS map, $\mathcal{L}_{\text{map}} \in \mathbb{R}^{H' \times W'}$, to strictly evaluate the covisible region:
\begin{equation}
    \text{LPIPS}_{\text{masked}} = \frac{1}{|\Omega'|} \sum_{p' \in \Omega'} \mathcal{L}_{\text{map}}(p'),
\end{equation}
where $\Omega'$ represents the mask $M$ downsampled to match the feature map resolution of the LPIPS network. We report LPIPS $\times 100$. Lower scores indicate better perceptual reconstruction.

\subsection{Semantic and Identity Alignment: DINOv3 \& SigLIP 2}
To evaluate whether the correspondence preserves the semantic identity and high-level details of the subject (crucial for our "Identity-Preserving Consistency" protocol), we employ deep feature similarity metrics using state-of-the-art vision foundation models.

\paragraph{DINOv3 Score.}
We utilize the {DINOv3} model~\cite{simeoni2025dinov3}. We extract the patch-level features from the last hidden state. To avoid artifacts from register tokens, we discard the first 5 tokens (CLS + registers) and reshape the remaining sequence into a spatial grid. The score is calculated as the cosine similarity between the warped image features $F_{\hat{I}_b}$ and target features $F_{I_b}$ averaged over the valid mask patches:
\begin{equation}
    \text{Sim}_{\text{DINO}} = \frac{1}{|P|} \sum_{k \in P} \text{Cos}(F_{\hat{I}_b}^{(k)}, F_{I_b}^{(k)}),
\end{equation}
where $P$ is the set of patches overlapping with the mask $M$.

\paragraph{SigLIP 2 Score.}
Complementary to DINO, we use {SigLIP 2 Large}~\cite{tschannen2025siglip} to capture multimodal semantic alignment. Similar to the DINO protocol, we extract the dense feature map from the vision tower's last hidden state, normalize the feature vectors, and compute the average cosine similarity over the masked region.

Higher DINOv3 and SigLIP~2 similarities ($\times100$) indicate closer feature alignment between the warped source and target. We interpret these scores together with the pixel-level and perceptual metrics, assessing whether the matching improvements are consistent across metrics.

\subsection{Implementation Details}
All metrics are implemented in PyTorch. For feature extraction (DINOv3, SigLIP 2), input images are resized to the model's native resolution (e.g., derived from patch size constraints) using bilinear interpolation, and masks are resized using nearest-neighbor interpolation.
\begin{itemize}
    \item \textbf{Masking Strategy:} A pixel/patch is considered valid if the mask value $> 0.5$ (or $> 128$ in 8-bit depth).
    \item \textbf{Normalization:} LPIPS inputs are normalized to $[-1, 1]$. DINOv3 and SigLIP 2 inputs follow the specific mean/std normalization required by their respective processors.
    \item \textbf{Evaluation Subset (IEG-Bench):} The quantitative results in Table~\ref{tab:baseline_IEG} are computed on a curated subset of approximately 100 image pairs selected for their geometric complexity and diverse content, providing a stress test for matching algorithms.
\end{itemize}

\subsection{Complementary Metrics and Paired Statistics}
\label{app:independent_metrics}

\paragraph{Reference-mask provenance.}
The reference masks are AI-assisted human annotations. FreeMatching predictions provide initial candidates, which annotators manually screen, annotate, and verify before evaluation. The finalized masks are then frozen and applied identically to all methods, rather than recomputed from each evaluated model's predictions. Full-MSE and whole-image OpenCLIP are evaluated without these masks.

For the 10k adaptation comparison in Table~\ref{tab:adapted_ieg}, each method is evaluated on the same image pairs. Ref-MSE uses a shared frozen reference mask in the RGB MSE formula above; Full-MSE replaces that mask with the full image. Ref-DINOv2 uses DINOv2 ViT-L/14~\cite{oquab2023dinov2} patch features at $280\times504$ resolution. We average patch cosine similarities with weights given by reference-mask coverage, retaining patches with coverage at least 0.25. OpenCLIP~\cite{cherti2023scaling} uses the LAION-2B-pretrained ViT-H/14 image encoder on $224\times224$ images and computes the cosine similarity between global image embeddings without masking. Both similarities are multiplied by 100. These metrics evaluate alignment with feature encoders distinct from the DINOv3 features used by FreeMatching.

We compute paired Full-MSE differences on the common evaluation pairs from IEG-Bench, sampling pairs with replacement for 10,000 bootstrap replicates (seed 20260728). The 2.5th and 97.5th percentiles give the reported 95\% confidence intervals. Positive advantages denote lower error for FreeMatching. Its image-level win rates on Full-MSE, Ref-DINOv2, and OpenCLIP are 68\%, 95\%, and 91\% against UFM, and 73\%, 78\%, and 87\% against RoMa.

\subsection{Weak Adaptation and Refinement-Domain Controls}
\label{app:refinement_controls}

Table~\ref{tab:adapted_classical} reports classical-task EPE for the three 10k-IEG-adapted methods. Table~\ref{tab:refinement_domains} compares FreeMatching-S1 with two refinement data sources: previously unseen, unlabeled Spring/FlyingThings pairs for 10k-Flow and IEG pairs for 10k-IEG. The flow-domain control uses no ground-truth flow during refinement. Each row reports one fixed checkpoint across the benchmarks.

\begin{table}[h]
\centering
\small
\caption{Classical-task EPE after 10k IEG adaptation. Lower is better; best results within this comparison are bold.}
\label{tab:adapted_classical}
\setlength{\tabcolsep}{4pt}
\begin{tabular}{@{}lccccc@{}}
\toprule
Method & Sintel Clean & Sintel Final & KITTI & ScanNet & ETH3D \\
\midrule
FreeMatching (10k-IEG) & 1.05 & \textbf{1.36} & 4.95 & \textbf{8.56} & \textbf{10.23} \\
UFM (10k-IEG)~\cite{zhang2025ufm} & \textbf{0.92} & 1.48 & \textbf{4.75} & 12.03 & 12.98 \\
RoMa (10k-IEG)~\cite{edstedt2024roma} & 1.24 & 2.49 & 5.91 & 12.03 & 11.26 \\
\bottomrule
\end{tabular}
\end{table}

\begin{table}[h]
\centering
\small
\caption{Effect of the refinement data source. We report IEG-Bench Ref-MSE ($\times100$) and EPE on classical benchmarks. Lower is better; best results within this comparison are bold.}
\label{tab:refinement_domains}
\setlength{\tabcolsep}{3pt}
\begin{tabular}{@{}lcccccc@{}}
\toprule
Checkpoint & Ref-MSE & Sintel Clean & Sintel Final & KITTI & ScanNet & ETH3D \\
\midrule
FreeMatching-S1 & 8.08 & 1.11 & 1.73 & 6.26 & 8.85 & 11.68 \\
FreeMatching (10k-Flow) & 7.26 & \textbf{0.98} & 1.59 & 6.46 & 8.91 & 11.64 \\
FreeMatching (10k-IEG) & \textbf{6.22} & 1.05 & \textbf{1.36} & \textbf{4.95} & \textbf{8.56} & \textbf{10.23} \\
\bottomrule
\end{tabular}
\end{table}

\section{Human Evaluation}
\label{app:human_eval}

To rigorously assess the perceptual quality of long-range correspondence and the effectiveness of our reconstruction consistency metric, we conducted a user study on the IEG-Bench dataset. While objective metrics provide scalable quantitative insights, human judgment remains the gold standard for evaluating semantic alignment and visual consistency. We compared FreeMatching with three representative baselines and recruited ten independent evaluators to rate the results blindly.

\begin{table}[h]
\centering
\small
\caption{\textbf{Quantitative comparison on IEG-Bench.} We report the Human Mean Opinion Score (MOS) and objective metrics. \textbf{Human (MOS)} is rated on a scale of 0-3 (higher is better). {LPIPS}~\cite{zhang2018perceptual} and {DINOv3}~\cite{simeoni2025dinov3} are scaled by 100. Our method achieves the highest ratings across all metrics. Best results are highlighted in \textbf{bold}.}
\label{tab:human}
\begin{tabular}{@{}lccc@{}}
    \toprule
    {Model} & {Human} $\uparrow$ & {LPIPS} $\downarrow$ & {DINOv3} $\uparrow$ \\
    \midrule
    SEA-RAFT~\cite{wang2024sea} & 0.49 & 72.45 & 58.14 \\
    RoMa~\cite{edstedt2024roma} & 1.39 & 48.91 & 70.15 \\
    UFM~\cite{zhang2025ufm} & 1.13 & 53.05 & 66.77 \\
    \midrule
    {FreeMatching (Ours)} & \textbf{2.25} & \textbf{45.41} & \textbf{80.01} \\
    \bottomrule
\end{tabular}
\end{table}

\noindent\textbf{Protocol and Criteria.}
The evaluation followed a blind protocol where method identities were anonymized and display orders were randomized. Evaluators were instructed to rate the alignment quality based on semantic consistency and texture preservation using a discrete 4-point scale: 
\begin{itemize}
    \item \textbf{0 (Failure)} for significant misalignment or severe distortion;
    \item \textbf{1 (Poor)} for visible errors affecting the main subject;
    \item \textbf{2 (Good)} for generally correct alignment with minor artifacts;
    \item \textbf{3 (Perfect)} for high-fidelity reconstruction indistinguishable from the target context.
\end{itemize}

\noindent\textbf{Results and Analysis.}
As reported in Tab.~\ref{tab:human}, FreeMatching achieves the highest MOS of \textbf{2.25}, significantly outperforming the strongest baseline, RoMa (1.39), and demonstrating substantial improvement over dense optical flow methods like SEA-RAFT (0.49).

Furthermore, to validate the reliability of our automated evaluation pipeline, we analyzed the correlation between human ratings and objective DINOv3 scores. We observe a strong positive Pearson correlation coefficient ($r=0.738$). This high correlation indicates that the DINOv3 metric aligns closely with human perception of correspondence quality, suggesting that our objective metrics serve as a reliable and efficient proxy for labor-intensive human evaluation in long-range alignment tasks.

\section{Beyond One-to-One Correspondence}
\label{app:one_to_many}

In IEG tasks involving object replication or synthesis, a query may admit multiple plausible correspondences. We outline a probabilistic extension of our FLUX-based framework to model the conditional distribution $p(C|I_a,I_b)$, providing a formulation for representing this ambiguity.

\paragraph{Training}
The extension replaces single-step regression with an iterative denoising objective. Given correspondence targets $C_{gt}$, we define noisy states $C_t$ according to a noise schedule. The model $f_\theta$, conditioned on image features and time $t$, can then be trained to recover clean coordinates directly ($x$-prediction):
\begin{equation}
\mathcal{L}_{\text{diff}} = \mathbb{E}_{t, \epsilon}  \mathcal{L}_{\text{Huber}} (f_\theta(I_a, I_b, C_t, t), C_{gt}) 
\end{equation}
Conditioning on the noisy correspondence state provides a mechanism for representing different matching hypotheses for the same image pair.

\paragraph{Sampling}
The corresponding sampling procedure starts from $C_T\sim\mathcal{N}(0,I)$ and iteratively refines the predicted clean correspondence. Different initial noise samples would allow the model to represent alternative matching hypotheses for the same image pair. Evaluating the validity and diversity of these hypotheses requires a dedicated experimental study.

This appendix describes a modeling extension; all reported experiments use the efficient single-pass regression model.

\section{Semantic Matching}
\label{app:semantic_matching}

Semantic correspondence commonly studies alignment across different instances of a category, whereas our task emphasizes the identity and local details of the same instance across IEG image pairs. SD-DINO~\cite{zhang2023tale} combines Stable Diffusion and DINOv2 features for zero-shot semantic correspondence.

\begin{figure}[t]
    \centering
    \includegraphics[width=\linewidth]{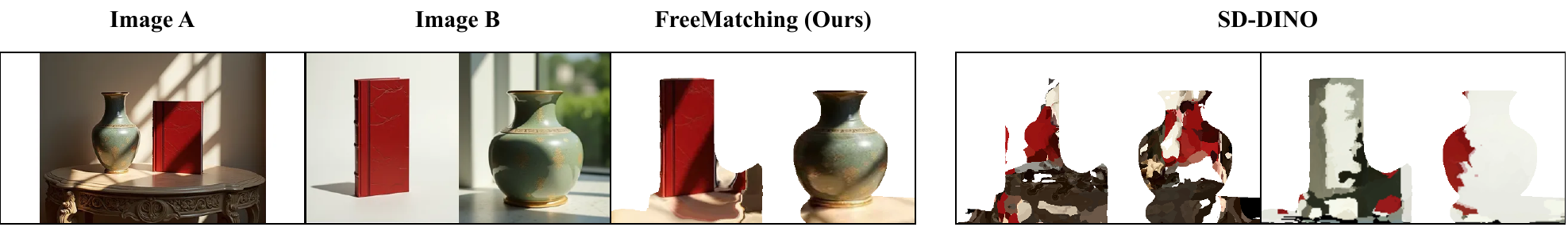}
    \caption{{Comparison with Semantic Matching.} In this multi-object example, the displayed SD-DINO~\cite{zhang2023tale} reconstructions, obtained with the prompts ``book'' and ``vase'', exhibit incomplete shapes and mixed object appearance. FreeMatching preserves more complete object shapes and local details without text prompts.}
    \label{fig:sem_match}
\end{figure}

Figure~\ref{fig:sem_match} compares the methods on a multi-object IEG example. The displayed SD-DINO results use the prompts ``book'' and ``vase'' in our evaluation configuration. Its reconstructions omit parts of the objects and mix their appearance, while FreeMatching preserves more complete object shapes and local details using only the image pair.

\end{document}